\documentclass{article}
\usepackage[utf8]{inputenc}
\usepackage{xcolor}
\usepackage{float}
\usepackage[justification=centering]{caption}
\usepackage[margin=0.5in]{geometry}
\usepackage[pdftex]{graphicx}
\graphicspath{{\Documents}}     
\DeclareGraphicsExtensions{.pdf,.jpeg,.png}
\usepackage{subfigure}
\usepackage{titlesec}
\usepackage{amssymb,amsmath}
\usepackage{moreverb}
\usepackage{multirow}
\usepackage[utf8]{inputenc}
\usepackage{multicol}

\title{\textbf{A Functional approach for Two Way Dimension Reduction in Time Series}}

\begin{multicols}{2}
\author{Aniruddha Rajendra Rao\\
Industrial AI Lab, Hitachi America, Ltd. R$\&$D\\
Santa Clara, CA\\
\and
Haiyan Wang\\
Industrial AI Lab, Hitachi America, Ltd. R$\&$D\\
Santa Clara, CA\\
\and
Chetan Gupta\\
Industrial AI Lab, Hitachi America, Ltd. R$\&$D\\
Santa Clara, CA}
\end{multicols}

\date{}

\begin{document}

\maketitle
\vspace{0.5cm}
\hspace{4.20cm}\{Aniruddha.Rao, Haiyan.Wang, Chetan.Gupta\}@hal.hitachi.com\\

\hspace{0.35cm}\textbf{Keywords:}   Functional Data Analysis, Deep Learning, Autoencoder, Functional Neural Network, Time series, Dimension Reduction.

\abstract{The rise in data has led to the need for dimension reduction techniques, especially in the area of non-scalar variables, including time series, natural language processing, and computer vision. In this paper, we specifically investigate  dimension reduction for time series through functional data analysis. Current methods for dimension reduction in functional data are functional principal component analysis and functional autoencoders, which are limited to linear mappings or scalar representations for the time series, which is inefficient. In real data applications, the nature of the data is much more complex. We propose a non-linear function-on-function approach, which consists of a functional encoder and a functional decoder, that uses continuous hidden layers consisting of continuous neurons to learn the structure inherent in functional data, which addresses the aforementioned concerns in the existing approaches. Our approach gives a low dimension latent representation by reducing the number of functional features as well as the timepoints at which the functions are observed. The effectiveness of the proposed model is demonstrated through multiple simulations and real data examples.}

\section{Introduction}\label{sec1}

Nowadays, with the rapid advancements in information technology, data are continuously collected at an astonishing frequency in many fields,  such as sensors installed on industrial 
equipment, economic factors, individual health, and environmental exposures. The exponentially growing volume of data poses challenges for storage, transfer, and analysis. 

Dimension reduction plays a key role in solving this problem by reducing the data in a meaningful manner. The objective of dimension reduction is to mathematically reduce the dimensionality by projecting the data to a lower dimensional subspace with minimum loss of information. The learned mathematical mapping plays an effective role in not only sorting out which variables are important but also how they interact with each other. Dimension reduction also helps us to deal with the curse of dimensionality, ease the transfer of information, reduce storage requirements, and reduce computation time. The usefulness of dimension reduction makes it an active area of research. Particularly, in the last few years, building dimension reduction models have proven to be beneficial in many fields, like time series, text, image, and video. This has attracted the interests of researchers from the scientific community, especially where non-scalar types of data have become prevalent \cite{WANG2016232,DR1,DR2, FAE1,10.5555/3045118.3045209}. Unfortunately, there is still a lot to be done in dimension reduction for time series data.

The purpose of this paper is to consider the problem of dimension reduction in time series data that occurs frequently in many areas of interest in industry and research like economics, meteorology, finance, health, sensors, etc. Examples include the weekly temperature, hourly sensor data, daily stock returns, and monthly blood pressure readings of a patient. This is a difficult and interesting problem as time series consists of random observations of the same variables at different timestamps, and the information is not independent nor linear because of the intricate temporal correlations. Mathematically, we are building mapping from multiple chronologically measured numerical variables within a certain time interval $\mathcal{S}$ to a few chronologically measured numerical variables within the same time interval $\mathcal{S}$.


In this paper, we consider a general setting similar to an autoencoder where the input and output time series are the same and we learn the dimension reduced form of the time series using the latent space representation. The standard machine learning methods for dimension reduction, such as Principal component analysis (PCA) and Autoencoders (AE), are designed to work with data that are recorded at regularly spaced points and in the finite dimensional space. These methods have the following limitations: ignoring intricate temporal correlations, unable to capture complex relations, scalar representation of time series data \cite{DR3}.


Functional data analysis (FDA) \cite{ramsay1997functional, book1, FDA} considers this time series information as functions over a continuum that are intrinsically infinite dimensional. Functional principal component analysis (FPCA) \cite{ramsay1997functional, book1, FDA} is the functional counterpart to PCA from the multivariate setting. FPCA is a useful dimension reduction tool that frequently serves as a key component in many functional analysis \cite{chiou2014multivariate, happ2018multivariate}. It allows us to do unsupervised learning of low-dimensional representations of the time series data by considering the entire time series as individual functional samples. FPCA helps to overcome some of the limitations mentioned above but they learn a linear representation of the data and therefore suffer from under fitting when the underlying mapping is complex. There is also a sufficient dimension reduction branch in FDA \cite{bingli2017,bingli2022} that learns a low-dimensional latent representation of the data which is non-linear but they are supervised. This limits the applicability of these methods. Also, they either don't work or don't perform well in the case of multiple functional features. 

Deep learning approaches have become very popular in the past few years. They offer some of the state-of-the-art methods for learning nonlinear representations of multiple types of data \cite{47658, AE1}. More recently, deep learning in functional data \cite{rao2020spatio,wang2019multilayer,9378087, thind2020deep, AR2} has gained a lot of momentum. Functional Autoencoders (FAE) \cite{FAE1} was inspired by Rossi et al. \cite{rossi2002functional, rossi2005representation}, introduced dimension reduction for functional data using deep learning. FAE adapted AE to the functional setting. They showed that the FPCA is a special case of FAE under certain conditions. While FAE has proven to beat a lot of the state-of-the-art methods \cite{FAE1}, unfortunately, it does not capture the true nature of functional data as only some of the layers have weight functions and the low dimensional latent representation of the time series data is scalar, which is restrictive. So far, all the methods we discussed have some kind of structural limitation, and to the best of our knowledge, the current works tackle either decreasing the number of functional features or decreasing the number of timepoints at which these curves are observed but there is very limited work on solving both of them together.


In this paper, we formulate the dimension reduction for the time series problem from the functional data analysis perspective. We innovatively identify a general mathematical mapping between the functional features, based on which a non-linear approach for two-way dimension reduction is proposed. Specifically, we introduce Bi-Functional Autoencoders (BFAE), a novel generalization of traditional autoencoders to functional data settings. The contributions of this paper are summarized as follows:

\begin{itemize}
    \item We propose a general functional mapping from multivariate temporal variables to themselves that embraces Functional Auto-encoders (FAE) as a special case. 
    \item We propose a new model that generalizes the well known neural network autoencoders to the functional data setting while preserving the functional nature of the data to address dimension reduction in time series.
    \item We explain how our approach can capture non-linear relations while reducing the dimensions in two-ways (number of features and time points) and, therefore, called Bi-Functional Autoencoders (BFAE).
    \item We demonstrate the effectiveness of the proposed approach in learning low dimensional non-linear representation of the time series through simulation experiments and two real-world examples.  
\end{itemize}

\section{Preliminaries}\label{sec2}

\subsection{Notations and Prior Art}\label{sec2.1}

This paper aims to develop an approach to map and represent the multiple time series variables (features) in a compact and efficient dimension reduced manner by leveraging the temporal dependencies within and between the involved variables. We begin by introducing useful notations, discussing the prior arts and problem definition, before proceeding to give the details of Bi-Functional Autoencoders (BFAE). Let us assume that for the  $i^{th}$ ($i\in \{1,2,...,N\}$) independent subject, we have $R$ features, that are continuously recorded within a compact time interval $\mathcal{S}\subseteq \mathbf{R}$. In particular, the observed timepoints of the $r^{th}$ feature for the $i^{th}$ subject are given by a $M^{(i,r)}_s$-dimensional vector $\mathbf{S}^{(i,r)}=[S^{(i,r)}_{1},...,S^{(i,r)}_{j},...,S^{(i,r)}_{M^{(i,r)}_s}]^T$, with $M^{(i,r)}_s$ representing the number of timepoints in the time series and $S^{(i,r)}_{j} \in \mathcal{S}$ for $i=1,...,n; r=1,...,R; j=1,...,M^{(i,r)}_s$. The corresponding time series are denoted as $\mathbf{X}^{(i,r)}=[X^{(i,r)}_{1},...,X^{(i,r)}_{j},...,X^{(i,r)}_{M^{(i,r)}_s}]^T$. The subscript $M^{(i,r)}_s$ reflects the fact that the measuring timestamps may vary across features and subjects. For the applicability of prior arts like PCA and AE, the data needs to be regular time series, therefore, we assume the number of timepoints across all features and subject is $M$. The observed data can be denoted as $\{\mathbf{X}^{(i,1)}, ..., \mathbf{X}^{(i,R)}\}_{i=1}^N$. We then concatenate the $R$ temporal features as $\mathbf{{X}}^{(i)}=[\mathbf{{X}}^{{(i,1)}^T},...,\mathbf{{X}}^{{(i,R)}^T}]^T$. Given samples $\{\mathbf{{X}}^{(i)}\}_{i=1}^N$, the problem definition can be stated as an unsupervised nonlinear multi-dimensional functional representation learning problem, formally defined as follows:
\begin{equation} \label{formulation1}
\mathbf{{Z}}^{(i)} = F(\mathbf{{X}}^{(i)}) 
\end{equation}

where $\mathbf{Z}^{(i)}=\{\mathbf{Z}^{(i,1)}, ..., \mathbf{Z}^{(i,R')}\}_{i=1}^N$ is a latent representation of the time series $\mathbf{{X}}^{(i)}$ with $R'<R$.

This problem formulation has a few disadvantages. Biases may get introduced when performing data pre-processing for having the same number of timepoints for each time series and these widely used prior arts have their own limitations in solving the dimension reduction problems in time series data when they assume them to be scalar values. These multivariate approaches are unable to capture the temporal dependencies among $\mathbf{{X}}^{(i,r)}$, $r=1,...,R$.

\subsection{Functional Data Analysis and an alternate Formulation}\label{sec2.2}

In the paper so far, we have seen the dimension reduction problem from a multivariate time series point of view. Let us look at an alternative problem formulation using functional data analysis (FDA). In FDA, we learn from random functions which are dynamically varying data over a continuum. There is a lot of application for FDA in a wide variety of fields, like time series, sensor data, image, and spatial data \cite{ramsay2006functional,FDA}. We can use this ability of FDA for dealing with continuous underlying curves $X^{(i,r)}(s), s \in  \mathcal{S}$ that is observed at discrete timepoints along a time interval. In functional analysis, the input or output has to be functional curves if not both. For given functional features $\{X^{(i,1)}(s), ..., X^{(i,R)}(s), s\in \mathcal{S}\}_{i=1}^N$, our problem of finding a latent representation of the data can be given as follows:

\begin{equation} \label{formulation2}
\mathbf{{Z}}^{(i)} (t) = F(\mathbf{{X}}^{(i)} (s)). 
\end{equation}

where $\mathbf{Z}^{(i)}(t)=\{\mathbf{Z}^{(i,1)}(t), ..., \mathbf{Z}^{(i,R')}(t), t\in \mathcal{S}\}_{i=1}^N$ is a latent representation of the time series $\mathbf{{X}}^{(i)(s)}$ with $R'<R$ and each time series is observed at $M'$ timepoints ($M'\le M$).

The popularity of FDA in time series application is evident with the boom of recent interest in this area of research. The assumption of smoothness can deviate in FDA as shown in \cite{yao2005functional}, allowing us to directly apply different FDA techniques to time series data. Also, compared to the common sequential models (RNNs, LSTMs) which learn fixed parameters over time, FDA enables us to learn feature effects that change over the interval $\mathcal{S}$ and build more efficient models \cite{9378087}. In the functional field, we have FPCA \cite{DR4, DR5} which can encode the information in a low-dimensional latent space but the mapping is linear. This can be a limiting factor because, in many real-world applications, the multiple functional features can have complex relations not just among them but across different timepoints as well.

The growing interest in deep learning and FDA has resulted in multiple successive deep functional models \cite{rossi2002functional, wang2019multilayer, AR1} for the different analytical tasks. FAE \cite{FAE1} is a deep learning approach for dimension reduction, a generalization of the autoencoder from the vector space to the functional setting. In this network, only the first and last layer can accommodate functions with the help of functional neurons given by \cite{rossi2002functional, rossi2005representation}, and the rest of the network consists of traditional neurons. Because of this, FAE can only give a latent representation of functional data in the scalar form, which is limiting. It is very common for functional features to have complex relations that can't be embedded in a scalar form. Thus, the scalar representation of functional data is inadequate.

In the next section, we propose a non-linear function-on-function autoencoder that leverages the power of fully connected Neural Networks while keeping the functional identity of the time series.

\section{Proposed Bi-Functional Autoencoder Model}\label{sec3}

\subsection{Our approach}

\begin{figure*}[]
	\centering
	\includegraphics[width=185mm]{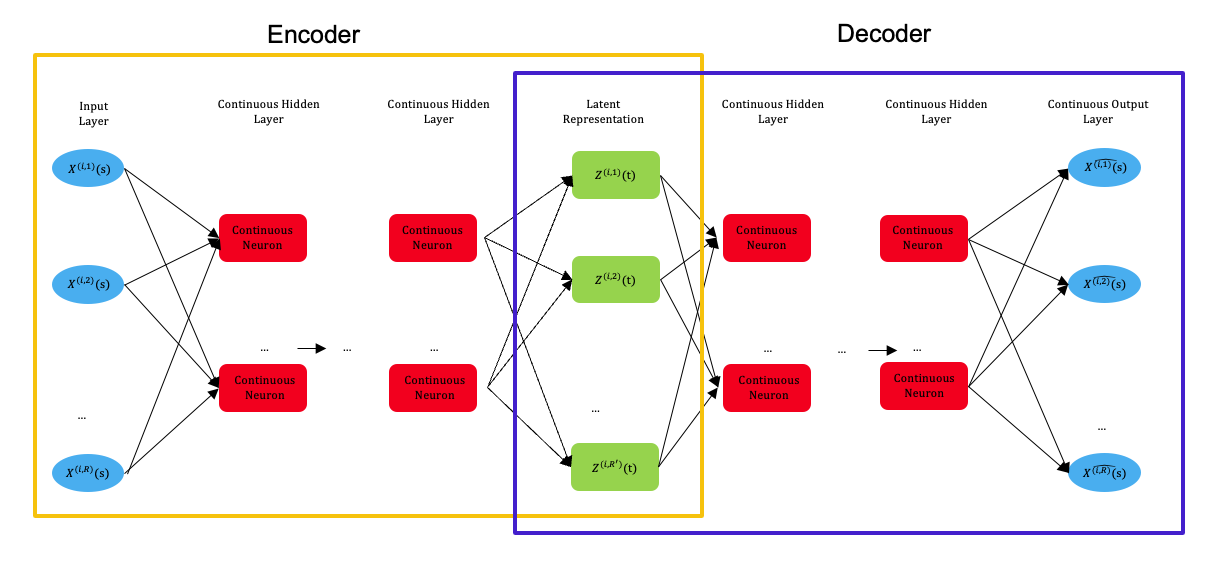} 
	\caption{General architecture for our proposed Bi-Functional Autoencoder}\label{Fig:arch}
\end{figure*}

The main goal of this paper is to learn a mapping $\mathbf{f}(\cdot)$ for the time series data to a low dimensional latent space as shown in Figure 1. To achieve this goal we make use of the idea of an autoencoder and Functional Neural Networks \cite{rossi2002functional, AR1, AR2}. An autoencoder (AE) is a popular dimension reduction technique which can learn non-linear latent presentation in a $R'$-dimensional space $\mathbf{R}^{R'}$ from a $R$-dimensional vector-valued input space  $\mathbf{R}^{R}$ (where $R'<R$) using the help of an encoder. The decoder part of the AE helps to map the learned latent representation back to the $R$-dimensional vector-valued input space  $\mathbf{R}^{R}$. The input and output of an AE are the same and we learn by measuring the reconstruction error of the output compared to the input. We want to use this idea but in a functional setting where we have an encoder that is mapping from a $R$-dimensional functional space to a $R'$-dimensional functional space and the decoder maps $R'$-dimensional functional space back to the $R$-dimensional functional space. We make use of continuous neurons defined in \cite{AR1, AR2} to enable us to achieve this task in a functional setting. The continuous neurons take in functional inputs and produce a functional output, thus leading to the preservation of the functional nature of the data.


The framework for our Bi-Functional Autoencoder (BFAE) Model, as seen in Figure 1, involves development of a continuous mapping from layer to layer by using multiple continuous hidden layers consisting of multiple continuous neurons as defined in \cite{AR1, AR2}. In Figure 1, we have three types of layers: an input layer, continuous hidden layers, and a continuous output layer. The input layer is the first layer that takes in the functional features, the continuous hidden layers consist of multiple continuous neurons, and the continuous output layer provides the reconstructed input functional features using the continuous neurons. The model learns with the help of functional weights that are continuous over time (or some other continuum). These functional weights consists of both univariate and bivariate functions, that is, we learn functions as well as surfaces. The key idea here is that we want to preserve the functional nature of the data throughout the whole reconstruction process across the network. This will give a richer structure to transform the data into low dimensional latent representation while exploiting the domain information and continuity of the functional features. We define the $l^{th}$ continuous hidden layer and its $r^{th}$ continuous neuron as:

\begin{equation}
\begin{split}
H^{(i,r)}_{(l)}(s)&=\sigma \Big(b^{(r)}_{(l)}(s) + \sum_{j=1}^{J} \int w^{(r,j)}_{(l)}(s,t)H^{(i,j)}_{(l-1)}(t)dt \Big) \label{e3}
\end{split}
\end{equation}
where $l=1,2,3,...,L$, $H^{(i,r)}_{(0)}(s)=X^{(i,r)}(s)$, $H^{(i,r)}_{(L)}(s)=\widehat{X^{(i,r)}(s)}$, $b^{(r)}_{(l)}(s) \in \mathcal{L}^2(\mathcal{S})$ is the unknown univariate intercept function, $w^{(r,j)}_{(l)} \in \mathcal L^2(\mathcal{S} \times \mathcal{S})$ is the bivariate parameter function for the $r^{th}$ continuous neuron in the $l^{th}$ continuous hidden layer coming from the $j^{th}$ continuous neuron of the $(l-1)^{th}$ continuous hidden layer and $\sigma(\cdot)$ is a standard activation function. The value of $J$, that can be considered as the number of features in each continuous hidden layer, is fixed for the input layer and the continuous output layer as $R$. We can tune $J$ to any scalar value for the rest of the continuous hidden layers. Similarly, the timepoints at which the output of the continuous neurons is observed is fixed for the input layer and the continuous output layer and can be changed for all the other continuous neurons in the remaining continuous hidden layers.

We will now look into the working of our network. We use the continuous neurons defined by Equation 3, to get from the input layer to a dimension reduced form of the input at an intermediate continuous hidden layer (say, it is the $l'^{th}$ layer). This is the encoder part of the BFAE approach called as the functional encoder. We can get the low dimensional representation of the inputs in the $l'^{th}$ continuous hidden layers as $\mathbf{Z}^{(i)}(s)=H^{(i)}_{(l')}(s)$ where $\mathbf{Z}^{(i)}(s)=\{\mathbf{Z}^{(i,1)}(s), ..., \mathbf{Z}^{(i,R')}(s), s\in \mathcal{S}\}_{i=1}^N$ and $H^{(i)}_{(l')}(s)=\{H^{(i,1)}_{(l')}(s), ..., H^{(i,R')}_{(l')}(s), s\in \mathcal{S}\}_{i=1}^N$, $R'<R$ and the function $\mathbf{Z}^{(i)}$ is observed at $M' \leq M$ timepoints. We move from this latent representation of the input towards the continuous output layer using the continuous neurons in the continuous hidden layers to get the reconstructed functional input values. This part of our approach is called the functional decoder.  Therefore, we have developed a mapping from a $R$-dimensional functional space to $R'$-dimensional functional space using the functional encoder, and the functional decoder maps $R'$-dimensional functional space back to the $R$-dimensional functional space. This latent representation $\mathbf{Z}^{(i)}(s)$ is very flexible in nature, where we can set the number of functional features as $R'$ and the number of timepoints observed as $M'$, according to the task in hand.

For simplicity, let us assume that Figure 1 has $R$ functional features in the input layer, $L$ continuous hidden layers, each with $J$ incoming and $R$ outgoing connections. BFAE is similar to an autoencoder in nature, but for functional data, we have to define the number of continuous hidden layers, the number of continuous neurons in each of the continuous hidden layers, and the number of timepoints at which these functions are observed for each continuous neuron. The activation function used in the continuous neurons is ReLU, tanh, sigmoid/logistic, or linear (continuous output layer). The forward propagation of the network is straightforward with the help of Equation 3. In the backpropagation phase, we learn using the functional surfaces (bivariate) and the functional intercepts (univariate) directly. We need the functional gradients to learn the functional parameters of the BFAE network. These functional gradients measure the change in a functional to a change in a function on which the functional depends.

We use the functional gradients given below to optimize our network's functional parameters. The optimization approach we use is the traditional gradient descent. The necessary assumptions and mathematical tools are discussed in \cite{rossi2002functional, wang2019multilayer, booko}. We use Fr\'echet derivatives, from the calculus of variation, for computing the functional gradients. The partial derivatives needed for the backpropagation are as follows:

\begin{equation} \label{e5}
\begin{split}
\frac{\partial H^{(i,r)}_{(l)}}{\partial b^{(r)}_{(l)}}(s)&= \sigma^\prime \Big(b^{(r)}_{(k)}(s) + \sum_{j=1}^{J} \int w^{(r,j)}_{(l)}(s,t)H^{(i,j)}_{(l-1)}(t)dt \Big)\\
\frac{\partial H^{(i,r)}_{(l)}}{\partial w^{(r,j)}_{(l)}}(s,t) &=\sigma^\prime \Big(b^{(r)}_{(l)}(s) + \sum_{j^\prime=1}^{J} \int w^{(r,j^\prime)}_{(l)}(s,t^\prime)H^{(i,j^\prime)}_{(l-1)}(t^\prime)dt^\prime \Big)\\
&\times \int \frac{\partial}{\partial w^{(r,j)}_{(l)}(s,t)} w^{(r,j)}_{(l)}(s,t)H^{(i,j)}_{(l-1)}(t)dt \\
&=\sigma^\prime \Big(b^{(r)}_{(l)}(s) + \sum_{j^\prime=1}^{J} \int w^{(r,j^\prime)}_{(l)}(s,t^\prime)H^{(i,j^\prime)}_{(l-1)}(t^\prime)dt^\prime \Big)\\
& \times H^{(i,j)}_{(l-1)}(t)\\
\end{split}
\end{equation}

where $l=1,2,3,...,L$, $H^{(i,r)}_{(0)}(s)=X^{(i,r)}(s)$, $H^{(i,r)}_{(L)}(s)=\widehat{X^{(i,r)}(s)}$ and $\sigma^\prime(\cdot)$ represents the first derivative of $\sigma(\cdot)$. In the backpropagation phase, we pass through the network backward from the continuous output layer to the input layer and calculate the partial derivatives of the continuous neurons with respect to the bivariate functional weight and the functional intercept. The loss function that we minimize is $\mathcal{L}(\theta)$, 
where $\theta$  is the collection of all functional parameters in the BFAE network. The loss function is given as follows:
\begin{equation} \label{lossmodel}
\mathcal{L}(\theta)={\frac{1}{N}  \sum_{i=1}^N \sum_{r=1}^R \int \Big(X^{(i,r)}(s)-\widehat{X^{(i,r)}(s)}\Big)^2 dt }
\end{equation}

The number of continuous hidden layers, the number of continuous neurons in each of the continuous hidden layers, and the number of observed timepoints for each continuous neuron can be considered as hyperparameters. We can also adjust our approach to accommodate irregular functional data or scalar variables. All these offer a lot of flexibility for our approach to work with different kinds of problems and adjust things according to the downstream process for different analytical tasks like prediction, classification, clustering, forecasting, and more.

\subsection{Connection of our approach to prior arts}\label{sec3.1}

Some of the current dimension reduction methods can be represented as a special case of BFAE. As we know, a single layer autoencoder (AE) with a linear activation function is analogous to PCA. In the same way, we can adjust the parameters of our model to act like FPCA. The FPCA is a special case of BFAE when our model has the following specifications: linear activation functions, a single continuous hidden layer, and the functional weights are constrained to be orthonormal. The Functional Autoencoder (FAE) \cite{FAE1} paper discusses how AE is a special case of FAE as FAE replaces the scalar weights and inner products of the AE with functional weights and inner products. In the same manner, FAE is a special case of our approach, BFAE. If we specify the functional encoder in our model to learn a scalar ($M'=1$) latent representation of the functional features, our approach essentially acts like FAE. Therefore, our approach is a generalization of most of the existing methods like PCA, FPCA, AE, and FAE.

\section{Numerical Experiments}\label{sec4}

In this section, we proceed to apply the proposed model to multiple simulation scenarios and two real-world problems. We compare our results against several state-of-the-art approaches, like PCA, AE, and FPCA, to show the effectiveness of BFAE. We demonstrate the following objectives through our results: 1) Our approach is effectively trained by the derived functional gradients 2) BFAE enables learning of the relation between the functional features 3) We show the efficacy of our approach by outperforming other methods.

\subsection{Simulations}

In the first experiment, we consider an individual predictor function (i.e. $R=1$) which is observed on a dense and regular grid. We generate $N=100, 1000$ iid random curves $\left\{X^{(1)}(t), \cdots, X^{(n)}(t)\right\}$ from a Gaussian process with mean $0$ and covariance given as follows:
\begin{equation}
  C_{X}(t, s)=\frac{\sigma^{2}}{\Gamma(\nu) 2^{\nu-1}}\left(\frac{\sqrt{2 \nu}|t-s|}{\rho}\right)^{\nu} K_{\nu}\left(\frac{\sqrt{2 \nu}|t-s|}{\rho}\right).
\end{equation}
where this is the Matérn covariance function and $K_{\nu}$ is the modified Bessel function of the second kind. The value of $\rho=0.5,$ $\nu=5 / 2$, and $\sigma^{2}=1 .$ These curves are realized at equally-spaced timepoint on the interval $[0,1]$ where $M=50, 250$ timepoints. We have specifically chosen a case where $N<M$ to check if it causes any issues with any of the approaches.

After generating the curves using a Gaussian process, we add some random noise  ($\epsilon_i(t)$) to them. The results are averaged over 100 simulations, and we divide the 100 samples into 80 for training and 20 for testing. The same ratio split is used for the case of 1000 samples. We report the Root Mean Square Error (RMSE) for the reconstruction of the functional features given by Equation (\ref{loss}) and in the case of the scalar approaches, we just report the standard RMSE ignoring the temporal aspect. For our approach in the first experiment, we consider multiple combinations of continuous hidden layers ($L$= 1, 3) and only one continuous neuron ($J=1$). The grid (s) value for latent representation in the continuous neurons is either $M$ or $M'=M/5$. We keep the network of AE similar to BFAE, and for PCA and FPCA we follow the $99\%$ variance of explained rule. Note that we can select $M'$ to be any scalar value in the range of ($1,M$) but we select this particular value ($M'=M/5$) to show the flexibility of our approach.

\begin{equation} \label{loss}
RMSE=\sqrt{\frac{1}{N}  \sum_{i=1}^N \sum_{r=1}^R \int \Big(X^{(i,r)}(s)-\widehat{X^{(i,r)}(s)}\Big)^2 dt }
\end{equation}




\begin{table}[]
\centering
\begin{tabular}{
c 
c 
c 
c 
c }
\hline
M         & \multicolumn{1}{c}{50}    & 250   & \multicolumn{1}{c}{50}    & 250   \\ \hline
Method/N & \multicolumn{2}{c}{100}           & \multicolumn{2}{c}{1000}          \\ \hline
PCA       & \multicolumn{1}{c}{0.483} & 0.418 & \multicolumn{1}{c}{0.469} & 0.398 \\ \hline
AE        & \multicolumn{1}{c}{0.409} & 0.443 & \multicolumn{1}{c}{0.390} & 0.406 \\ \hline
FPCA      & \multicolumn{1}{c}{0.390} & 0.351 & \multicolumn{1}{c}{0.356} & 0.343 \\ \hline
BFAE      & \multicolumn{1}{c}{0.301} & 0.265 & \multicolumn{1}{c}{0.102} & 0.098 \\ \hline
BFAE (M’) & \multicolumn{1}{c}{0.331} & 0.274 & \multicolumn{1}{c}{0.117} & 0.104 \\ \hline
\end{tabular}
\caption{Comparing RMSE of different methods for capturing a single time series function.}
\label{tab1}
\end{table}

We observe from Table I that PCA and AE perform similarly on the temporal curves. There is some gain in performance for FPCA but our approach performs the best irrespective of the sample size ($N$) and the number of timepoints ($M$). The difference in performance increases when the sample size and the number of timepoints increases, as our approach is a deep learning model, it learns better with more information. When we set the number of observed timepoints as $M'$, our approach loses some performance because we are restricting the true data to a smaller latent space. But, even under this scenario of representing the true curves using 10 or 50 timepoints rather than 50 or 250, we are still performing better than all the other approaches.

\begin{figure}[htbp]
	\centering
	\includegraphics[width=90mm]{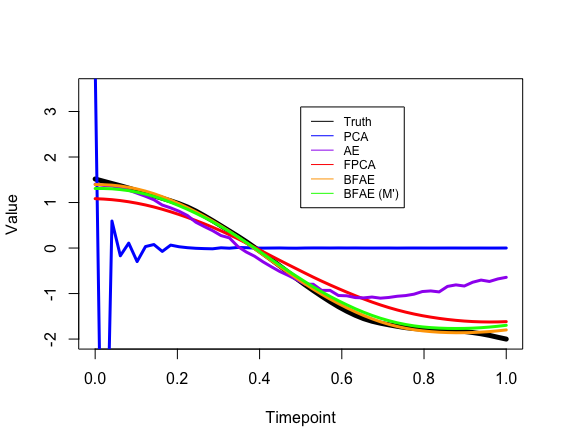} 
	\caption{Curve reconstruction comparison for different methods when $N=1000$, $R=1$, and $M=50$}\label{Fig:arch1}
\end{figure}

Figure 2 shows us the reconstructed curves for a random sample from different approaches for the simulation setting of $N=1000$, $R=1$, and $M=50$. We see a similar story compared to Table I where PCA is performing poorly, and the shape is very different from the truth. PCA is a linear approach and can't capture the temporal information in the data. AE and FPCA perform better than PCA by getting the general shape correct but are still far off from the truth. Our approach, BFAE, irrespective of the number of timepoints observed, is able to reconstruct the truth well and captures the shape of the curve correctly over the whole time interval.

We extend the first experiment to include more functional features. In the second experiment, we increase the number of features to $R=10$. We follow the same procedure to generate these 10 curves for the same set of values of $N$ and $M$. The results discussed below are an average of 100 simulations after adding random noise and using the same splits for training and testing. For our approach, we consider multiple combinations of continuous hidden layers ($L$= $1$, 3), continuous neurons ($J$= 1, 2, 4) and we set $R'=4$. The grid (s) value for latent representation in the continuous neurons is either $M$ or $M'=M/5$.

\begin{table}[]
\centering
\begin{tabular}{
c 
c 
c 
c 
c }
\hline
M         & \multicolumn{1}{c}{50}    & 250   & \multicolumn{1}{c}{50}    & 250   \\ \hline
Method/N  & \multicolumn{2}{c}{100}           & \multicolumn{2}{c}{1000}          \\ \hline
PCA       & \multicolumn{1}{c}{1.527} & 1.525 & \multicolumn{1}{c}{1.518} & 1.518 \\ \hline
AE        & \multicolumn{1}{c}{0.471} & 0.490 & \multicolumn{1}{c}{0.489} & 0.484 \\ \hline
FPCA      & \multicolumn{1}{c}{0.386} & 0.417 & \multicolumn{1}{c}{0.365} & 0.404 \\ \hline
BFAE      & \multicolumn{1}{c}{0.328} & 0.292 & \multicolumn{1}{c}{0.109} & 0.132 \\ \hline
BFAE (M’) & \multicolumn{1}{c}{0.359} & 0.314 & \multicolumn{1}{c}{0.147} & 0.169 \\ \hline
\end{tabular}
\caption{Comparing RMSE of different methods for capturing multiple ($R=10$) time series functions.}
\label{tab2}
\end{table}

Table II shows that the increase in the functional features results in a deterioration in performance for PCA. While the other two approaches perform better than PCA, our approach performs the best irrespective of the sample size ($N$) and the number of timepoints ($M$). The difference in performance again increases when the sample size and number of timepoints increases. We observe that the errors have increased marginally compared to the previous table as we have added more features. The behavior of decreasing the number of timepoints to $M'$ is similar as well.

\subsection{Real Data applications}\label{sec4.1}

We consider two real data sets. The first is the speech recognition data from TIMIT (available at http://statweb.stanford.edu/tibs/Ele-mStatLearn/) where we have speech signals for different phonemes. Audio information is available abundantly, but they are measured at a very high frequency requiring high storage capacity. A low dimensional latent representation of this data would be very useful in the industry. The other example deals with the relation between electricity demand and temperature in the city of Adelaide, Australia
. This data is important because of the high costs related to the storage of electricity and understanding the effect of temperature on demand can lead to information based operational steps.

For speech recognition data, we set up the experiment similar to \cite{bingli2017, AR1}, where we have voice signals ($R$=1) for two phonemes (response) transcribed as follows: “aa” as the vowel in “dark” and “ao” as the first vowel in “water”. We build a classifier model with the help of FLM after applying dimension reduction to the voice signals. Figure 3  shows that it is difficult to separate the two groups where we see the two phoneme curves computed by a log-periodogram of 150 ($M=150$). Each phonemes information is recorded over 150 points, reduction of such data is beneficial, especially with the potential to expand this low dimension representation to much higher voice signals. We have 800 functional samples, we split these samples into 640 samples for training and 160 samples for testing.

The Adelaide data (available in $fds$ package in R-software) has the temperature and electricity demand records from 7/6/1997 to 3/31/2007 (508 weeks) measured daily at half-hourly rate ($M=48$). We consider each day of the week as a feature ($R$=7) and try to map the relation between temperature and demand using Functional Linear model (FLM) \cite{book1,FDA, ramsay2006functional}. Before we use FDA to model this relation, we use BFAE to represent the temperature curves in a compact low dimensional form. We split this data into 400 samples for training and 108 samples for testing. Figure 4 shows the half-hourly temperature and electricity demand (Megawatts) for the whole 508 weeks. We can see from the image that the temperature curve for each day of the week follows a similar pattern and learning that pattern can lead to a reduction in dimension with minimum loss of information.

\begin{table}[]
\centering
\begin{tabular}{
c 
c 
c }
\hline
Method/Data & Phonemes & Adelaide \\ \hline
PCA         & 14.960   & 18.633   \\ \hline
AE          & 1.790    & 2.354    \\ \hline
FPCA        & 1.242    & 0.636    \\ \hline
BFAE        & 1.126    & 0.581    \\ \hline
BFAE   (M’) & 1.128    & 0.615    \\ \hline
\end{tabular}
\caption{Comparing RMSE of different methods for capturing prediction functions for the real data.}
\label{tab3}
\end{table}

We can observe from Table III that PCA again is having difficulties in reducing the information for both the real data sets. While AE and FPCA are performing better than PCA, BFAE performs the best. For phonemes data, we even reduced the points to $M'=30$ and still BFAE performs better than other approaches. The difference between BFAE with timepoints as $M$ and $M'$ is negligible, indicating that the low dimension representation is very rich in information. While for Adelaide data, we reduce the number of features to 4 ($R'=4$) and the timepoints to $M'=12$ and produce the best results using our approach with timepoints as $M$. Our intuition to reduce the temperature feature is certainly validated as seen from Figure 4 and Table III.

Modeling results can be seen in Table IV where we build a classifier for the phonemes data, to predict if the voice signal is for “aa” or “ao”. For the Adelaide data, we map the temperature information for the 7 days of the week to predict the electricity demand for the 7 days of that same week. Both these models are built with the help of the original data and the low dimension latent representation learned using BFAE. We ignore the other approaches as the goal here is to demonstrate that our approach gives modeling results at least as good as the original data by capturing the signals. We can see from Table IV that our approach not only does a good job of modeling but also performs as well as the original data and has a lower tendency to over fit as it contains less noise. The error increases in both cases as we reduce the timepoints to $M'$ but the performance is still competitive.


\begin{table}[]
\centering
\begin{tabular}{
c 
c 
c 
c 
c }
\hline
Real Data                 & Data  & Original & BFAE    & BFAE  ($M'$) \\ \hline
                                                                                                  & train & 0.175    & 0.185   & 0.205      \\ \cline{2-5} 
\multirow{-2}{*}{\begin{tabular}[c]{@{}c@{}}Phonemes\\    \\ \end{tabular}} & test  & 0.200    & 0.190   & 0.210      \\ \hline
                                                                                                  & train & 188.363 & 164.105 & 199.758    \\ \cline{2-5} 
\multirow{-2}{*}{\begin{tabular}[c]{@{}c@{}}Adelaide\\    \\\end{tabular}}                 & test  & 220.739  & 184.518 & 215.294    \\ \hline
\end{tabular}
\caption{Comparison of errors (Classification error for Phonemes and RMSE for Adelaide) of different methods for modeling the response.}
\label{tab3.1}
\end{table}

Overall, our approach does the best job in reducing the information into a low dimensional latent space while still maintaining a competitive performance at modeling different tasks compared to the original data. We are able to represent the information in a compact data rich manner and are also able to reconstruct it back to the original scale without much loss of information.



\begin{figure}[]

		\centering
		\includegraphics[width=9.00cm]{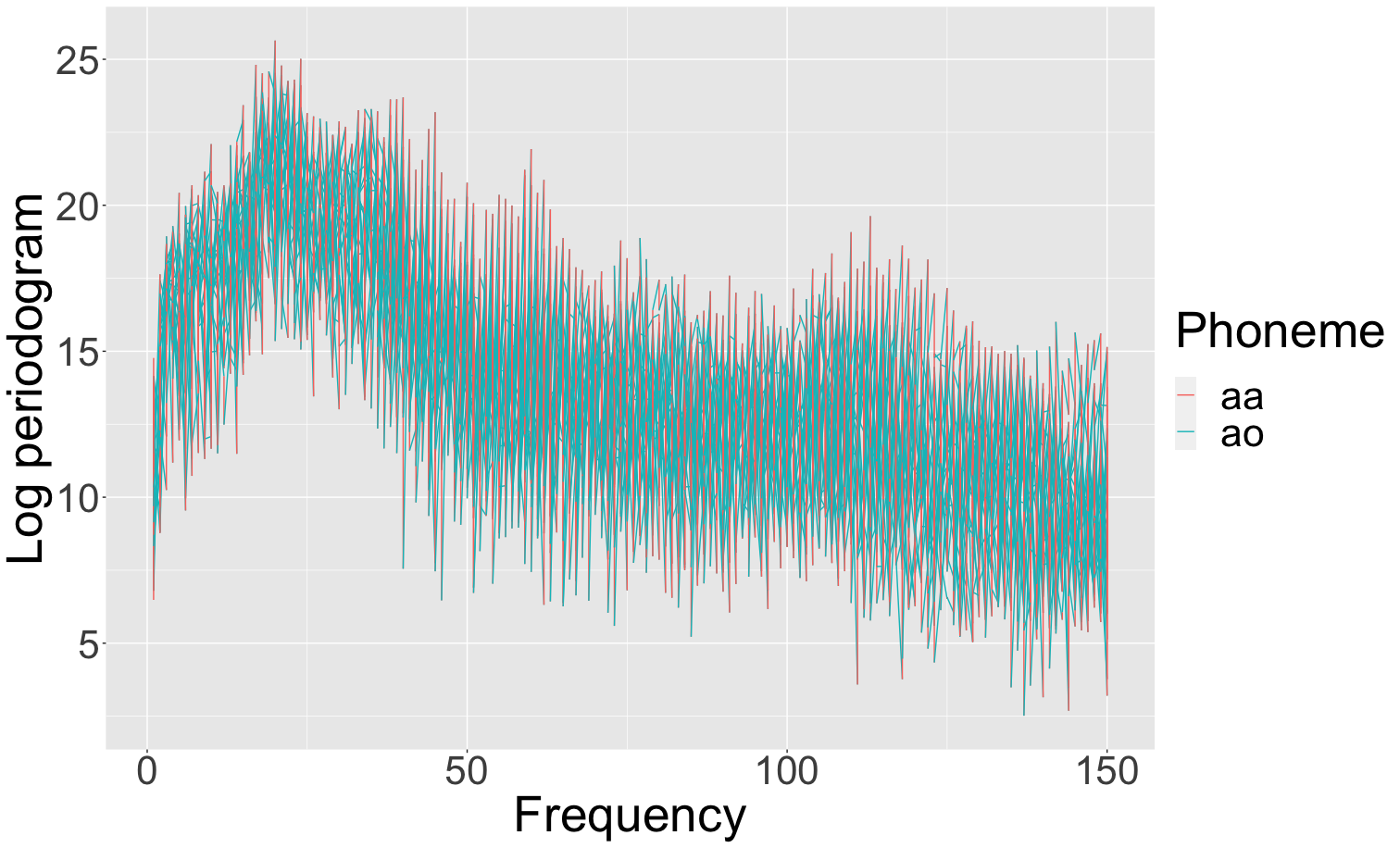}

	\caption{ Voice signal curves for the Phonemes data}\label{dataset1}
\end{figure}

\begin{figure}[]
	\centering
	\begin{subfigure}
		\centering
		\includegraphics[width=16.55cm,height=3.00cm]{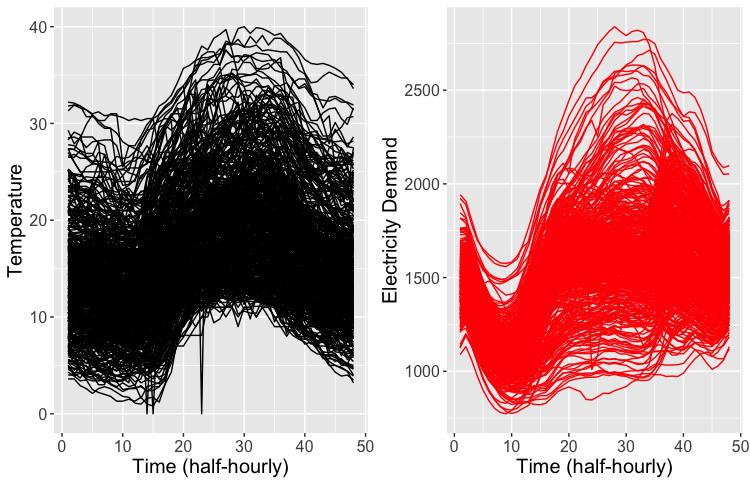}		
	\end{subfigure}
    \begin{subfigure}
		\centering
		\includegraphics[width=16.55cm,height=3.00cm]{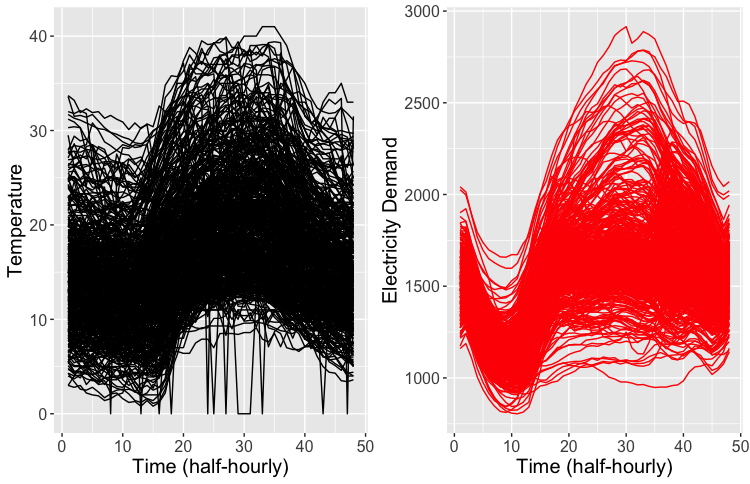}		
	\end{subfigure}

	\begin{subfigure}
		\centering
		\includegraphics[width=16.55cm,height=3.00cm]{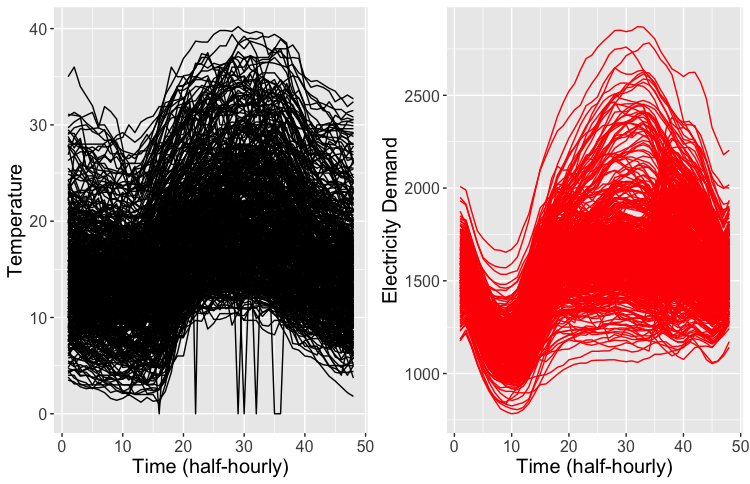}		
	\end{subfigure}
	\begin{subfigure}
		\centering
		\includegraphics[width=16.55cm,height=3.00cm]{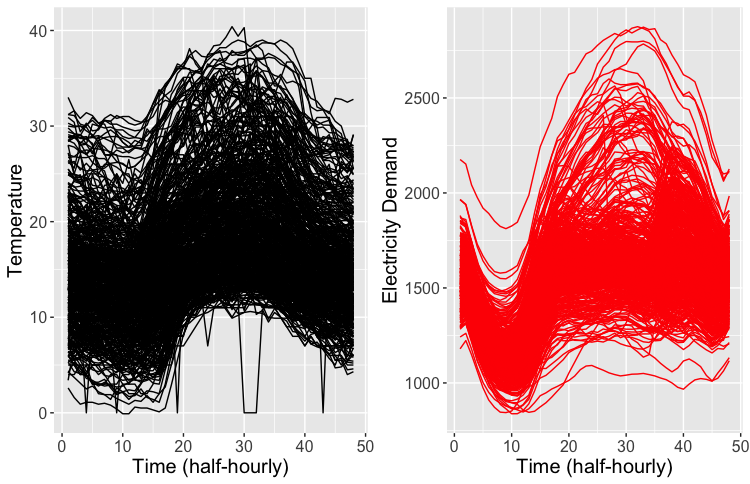}		
	\end{subfigure}
	\begin{subfigure}
		\centering
		\includegraphics[width=16.55cm,height=3.00cm]{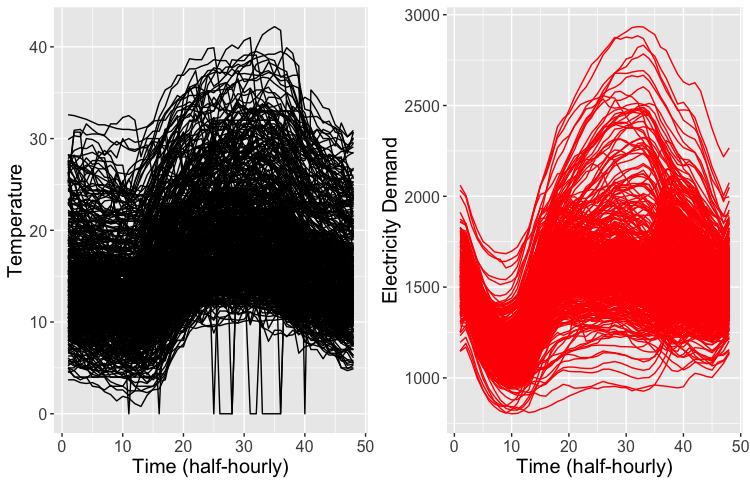}		
	\end{subfigure}
	\begin{subfigure}
		\centering
		\includegraphics[width=16.55cm,height=3.00cm]{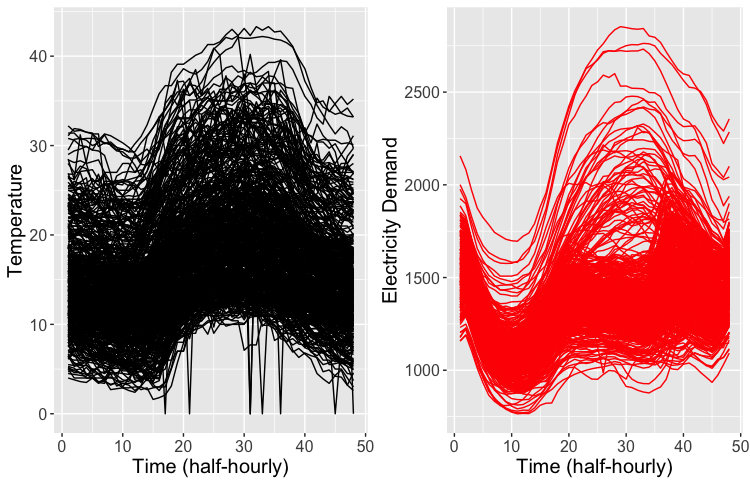}		
	\end{subfigure}
	\begin{subfigure}
		\centering
		\includegraphics[width=16.55cm,height=3.00cm]{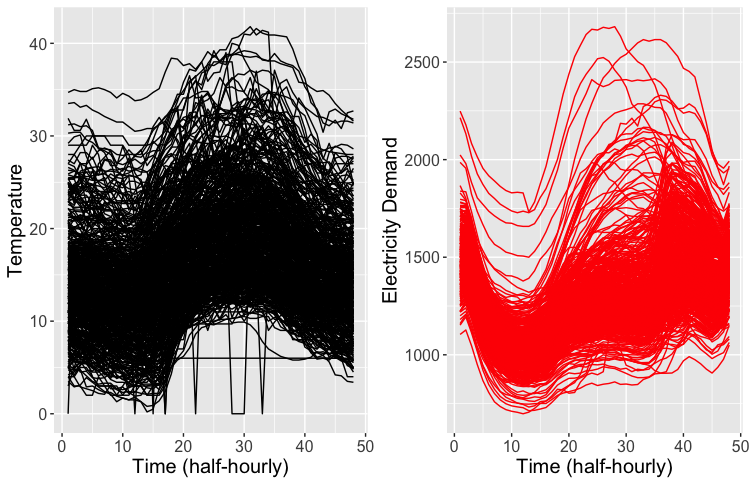}		
	\end{subfigure}


	\caption{ Day curves (Monday to Sunday, top to Bottom) of temperature and electricity demand for Adelaide}\label{dataset2}
\end{figure}




\section{Conclusions}\label{sec5}


In this paper, we proposed a novel model for multivariate time series dimension reduction. Our approach helps to perform two-way dimension reduction by reducing the number of features and the number of timepoints at which the time series is observed. The proposed Bi-Functional Autoencoder (BFAE) reduces the input into a low dimension latent representation using a functional encoder and reconstructs the information back using a functional decoder. Our proposed approach has a lot of advantages compared to current methods, including its ability to  represent the time series data in a flexible manner, capture timely varying correlations among features, ability to deal with different kinds of data, and capture non linear relations. Along with superior simulation results, we showed the proficiency of our approach in two real-world examples in comparison with several common practices in the prior art. Our approach produced smaller errors in reconstruction and modeled the data as well as the original information. We expect the proposed model to be widely used in diverse real-world problems where the goals are to reduce the transfer load of huge amount of time series data, store a large amount of time series information effectively, and reduce the computation cost of different analytical tasks while retaining the performance.


\bibliographystyle{plain}
\bibliography{arxiv_paper}

\end{document}